\documentclass[sigconf,screen,nonacm]{acmart}

\AtBeginDocument{%
  \providecommand\BibTeX{{%
    \normalfont B\kern-0.5em{\scshape i\kern-0.25em b}\kern-0.8em\TeX}}}

\setcopyright{acmcopyright}
\copyrightyear{2021}
\acmYear{2021}
\acmDOI{10.1145/xxxxxxx.xxxxxxx}

\DeclareMathOperator*{\argmax}{arg\,max}

\begin{document}

\title{3D Parametric Wireframe Extraction Based on Distance Fields}

\author{Albert Matveev}
\email{albert.matveev@skoltech.ru}
\affiliation{%
  \institution{Skolkovo Institute of Science and Technology}
  \city{Moscow}
  \country{Russia}
}

\author{Alexey Artemov}
\affiliation{%
  \institution{Skolkovo Institute of Science and Technology}
  \city{Moscow}
  \country{Russia}
}

\author{Denis Zorin}
\affiliation{%
  \institution{New York University}
  \city{New York}
  \country{USA}
}
\affiliation{%
  \institution{Skolkovo Institute of Science and Technology}
  \city{Moscow}
  \country{Russia}
}

\author{Evgeny Burnaev}
\affiliation{%
  \institution{Skolkovo Institute of Science and Technology}
  \city{Moscow}
  \country{Russia}
}

\begin{abstract}
  We present a pipeline for parametric wireframe extraction from densely sampled point clouds. Our approach processes a scalar distance field that represents proximity to the nearest sharp feature curve. In intermediate stages, it detects corners, constructs curve segmentation, and builds a topological graph fitted to the wireframe. As an output, we produce parametric spline curves that can be edited and sampled arbitrarily. We evaluate our method on 50 complex 3D shapes and compare it to the novel deep learning-based technique, demonstrating superior quality. 
\end{abstract}




\begin{teaserfigure}
  \centering
  \includegraphics[width=\textwidth]{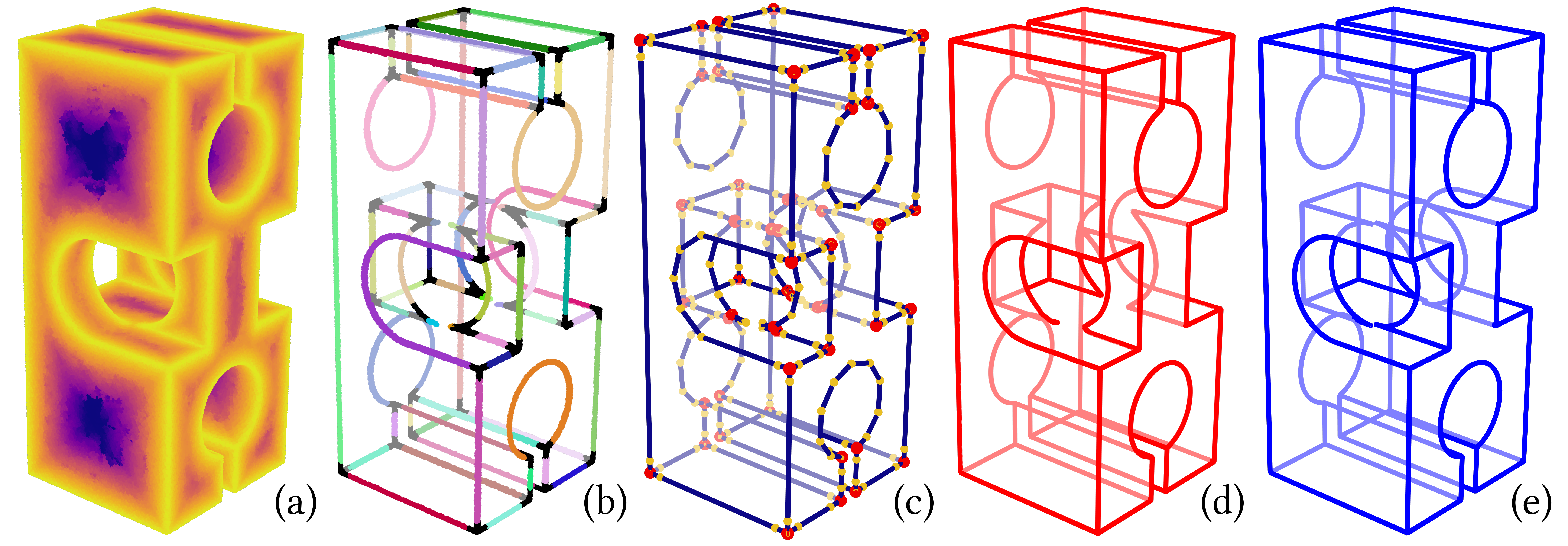}
  \caption{Parametric wireframe extraction pipeline: (a) -- dense point cloud with estimated distance field, (b) -- sharp point skeleton with color-coded segmentation into individual curves (note the black clusters corresponding to the detected corner neighborhoods), (c) -- topological graph with final corner points (red), (d) -- extracted parametric wireframe, (e) -- ground-truth parametric wireframe.}
  \label{fig:teaser}
\end{teaserfigure}

\maketitle

\section{Introduction}
\label{sec:intro}
In recent years, the importance of 3D data processing has increased considerably. The availability of a multitude of scanning devices capable of capturing the geometry of the scanned objects has led to a vast growth of accessible 3D datasets. This leads to the need to develop new and improve existing algorithms for more efficient data processing.

Conversion of the raw 3D scan data into high-order geometry representations is commonly referred to as a geometry processing pipeline. At each step of the pipeline, there are several possible problems to tackle, including point cloud registration~\cite{tam2012registration}, upsampling~\cite{huang2013edge}, estimation of various geometric properties (normals~\cite{osculatingjets}, curvature~\cite{kalogerakis2007robust}, directional fields~\cite{vaxman2016directional}), and surface reconstruction~\cite{kazhdan2006poisson} from unordered point sets. The range of possible applications is vast, including robotic navigation, self-driving vehicles, virtual reality, and computer graphics. 

The tasks mentioned above can benefit from geometric deep learning~\cite{guo2020deep}. However, until recently, it was hard to generate ground truth for most geometry processing problems. One of the most convenient data representations for such cases is CAD models that constitute a 3D volume as a set of parametric curves and surfaces. CAD shapes usually are reproducing mechanical parts or other human-made objects of complex geometry that are characterized by so-called sharp feature curves, where the normals are discontinuous across the curve. For computer graphics and computer-aided design, it is essential to reconstruct such curves as parametric entities. 

Recently, a new approach to sharp feature lines detection has occurred -- Deep Feature Estimators (DEFs)~\cite{matveev2020def}. In DEF, the authors introduced a novel learning-based framework to infer sharp feature curves in unordered 3D point clouds. Unlike the previously introduced neural networks, which are modeling sharp features as segmentation masks, DEF aims to regress a real-valued scalar distance field. We argue that distance fields are ideally suited for parametric curve extraction.

In this paper, we introduce a pipeline for processing sharp feature predictions into the 3D parametric splines. Our approach processes DEF output to base the inference and optimization procedures on the estimated distance-to-feature fields. The pipeline consists of the following steps: (1) a corner detection procedure that takes into account the per-point distance estimates; (2) curve clustering for curve instance segmentation; (3) construction of the topological graph approximating a sharp curves wireframe; (4) fitting spline curves to the corresponding point sets. We tested our approach on a set of 50 3D shapes sampled as point clouds and processed by DEF. We demonstrated that the described approach robustly detects individual curves and can fit an accurate parametric wireframe. Refer to Fig.~\ref{fig:teaser} for an overview of the whole pipeline.
\section{Related Work}
\label{sec:related}

\paragraph{Datasets.}
Several mesh-based and CAD datasets have become available lately. ShapeNet~\cite{shapenet} has two open subsets ShapeNetCore (51,300 models) and ShapeNetSem (12,000 models), that are widely used in many publications for classification and semantic segmentation problems. However, ShapeNet polygonal meshes are frequently of unsatisfactory quality. Another corpus is ModelNet~\cite{modelnet}, which includes 151,128 shapes. It is commonly utilized as a classification and normals estimation benchmark. These collections are mesh datasets and do not contain any geometric labeling. ABC dataset~\cite{koch2019abc}, comprising over a 1,000,000 high-quality CAD models, provides a rich annotation with geometric, topological, and semantic information. Specifically, the curves are stored as parametric entities, enabling us to assess the wireframe extraction performance quantitatively.

\paragraph{Sharp Feature Curves Estimation.} 
Detection of sharp edges has a wide range of applications; hence this topic has been studied extensively. However, all of the analytical methods are performing hard label segmentation of points, which is limiting. Typically, local differential quantities are estimated via local sets on $k$ nearest neighbors. One family of such approaches performs normals estimation with principal components analysis (PCA), then applies clustering to identify areas where normals change rapidly~\cite{weber2010sharp,fastrobustsharp,bazazian2015fast}. 
Another group of algorithms relies on the analysis of covariance matrices~\cite{merigot2010voronoi,dey2013voronoi}. Such covariance analysis methods commonly determine the ellipsoid shape constructed on the local set and then mark sharp edges by deriving per-point classification. Typically, extracted features are noisy and unstable, and these methods require careful per-model parameter tuning. Moreover, although hard labeling is helpful in various settings, it does not apply to curve inference due to imprecise detection techniques.

\paragraph{Deep Learning for Sharp Feature Detection.} 
A few learning-based approaches have been proposed for sharp feature detection. In~\cite{yu2018ec} sharp edges are applied to point cloud upsampling and are treated as a supplementary task with an additional loss term. Attention mechanism was adapted for more precise point cloud neighborhood selection in~\cite{matveev2020geometric}, which was beneficial for feature lines detection and normals estimation. The most recent architecture was presented in paper~\cite{Bazazian-EDCNet2021} in which a capsule neural network was used to construct a classifier that predicts edge and non-edge points in 3D point clouds. All of the approaches mentioned above, similarly to analytical methods, perform per-point classification. 
Recently, a new approach to sharp feature lines detection has occurred in DEF~\cite{matveev2020def} that can represent features with so-called distance fields. Our pipeline is based upon this paper since it provides a soft detection suitable for curve inference precisely on edge.

\begin{figure*}
\centerline{
\includegraphics[width=0.8\textwidth]{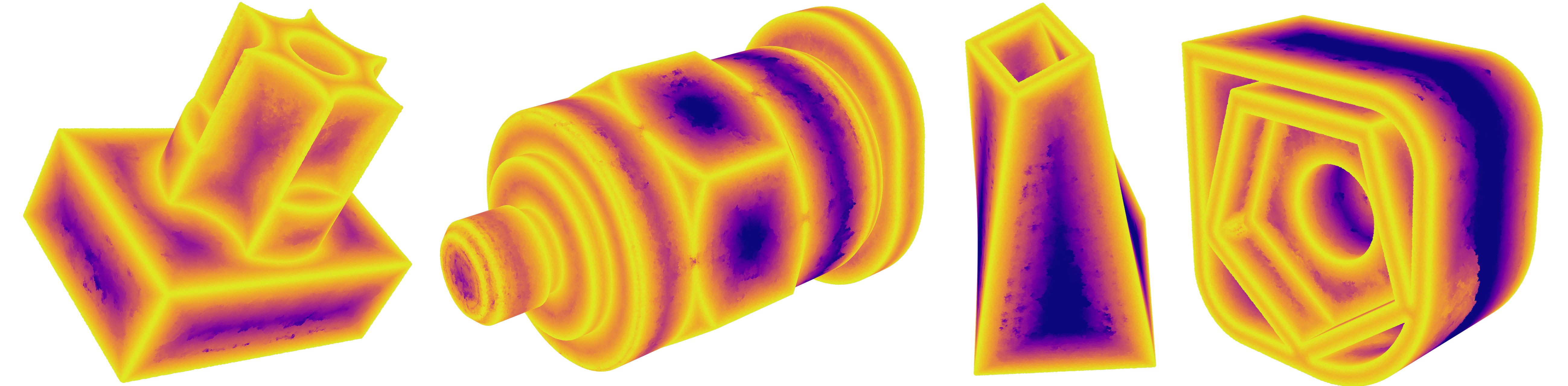}
}
\caption{Examples of DEF predictions: dense point clouds with estimated distance fields. Although predictions are not always smooth, sharp feature curves are detected robustly.}
\label{fig:def_examples}
\end{figure*}

\paragraph{Parametric Curve Extraction.}
The vast majority of the prior art on vectorization is applied to 2D data: sketches~\cite{favreau2016fidelity} and technical drawings~\cite{egiazarian2020deep}. Graph wireframe extraction was applied to landscape and interior modeling in~\cite{zhou2019end,zhou2019learning}. Such methods are not directly applicable to our setup. The most direct competitors are PIE-Net~\cite{wang2020pie} and PC2WF~\cite{Liu:2021:PC2WF}. PIE-Net utilizes a standard PointNet++ architecture to obtain the sharp edge segmentation and detect corners. After that, it derives parametric curves as splines. PC2WF approaches the problem similarly, attempting to infer the wireframe directly from detected corners. The main drawback, in this case, is that the only feasible type of curve is a straight segment. We argue that this is limiting; hence we conduct comparisons only against PIE-Net.

\section{3D Wireframe Extraction}
\label{sec:design}

\subsection{Description of Distance Field}
The distance-to-feature regression presented in DEF is constructed as follows. The point cloud (possibly, a real scan) is separated into several point cloud patches, representing a part of the whole 3D model. As an input, DEF accepts a point cloud patch $P = \{(x_i,y_i,z_i)\}_{i=0}^{4096}$ of 4,096 points. Such patches are separately processed with a neural network. 

For each point in the patch, the network outputs distance-to-feature estimates $\widehat{d}_i\in\lbrack0, 1\rbrack$ for each point. These distance-to-feature estimates indicate the distance from a chosen point to the closest sharp feature curve. Thus, the field's zero level set is the actual location of the feature line. Furthermore, such properties liberate from choosing the segmentation width and point cloud sampling issues. It is problematic while performing least-squares fitting to a point cloud, which is inevitably erroneous since point samples are not on the feature curve exactly. 

Once the distances are estimated, patches are combined to render the whole shape again, and the predicted distances are aggregated to ensure the global consistency of the field. This patch-based design of the approach facilitates processing large point sets, which are unfeasible for other methods. In addition, the high density of points results in more accurate solutions for downstream tasks. See the examples of point cloud shapes and the corresponding DEF predictions in Fig.~\ref{fig:def_examples}.

\subsection{Extracting Parametric Curves}
We denote a point cloud as $\mathcal{P} \in \mathbb{R}^{N \times 3}$, and a corresponding vector of predicted distance estimates as $\widehat d \in \mathbb{R}^N$. At the first stage, we select only points within a threshold distance $T_{\text{dist}}$ to the closest feature line, thus constructing a point cloud skeleton $\mathcal{P}_{\text{sharp}}$. Then, we detect corners in this point cloud skeleton (Sec.~\ref{sec:design:corners}). The reason for this is two-fold: first, corner detection helps with clustering the rest of the skeleton to obtain the curve segmentation described in Sec.~\ref{sec:design:seg}; second, we construct the topological graph as a set of segments connecting such corner points (Sec.~\ref{sec:design:graph}). Finally, we discuss spline fitting and optimization in Sec.~\ref{sec:design:splines}.

\subsubsection{Corner Detection}
\label{sec:design:corners}
Corner detection starts with selecting 10\% of points from the point cloud skeleton via farthest point sampling to serve as query points, which are centers of a set of balls $\mathcal{N}_i$ covering $\mathcal{P}_{\text{sharp}}$. The radius of balls $R_{\text{corner}}$ needs to be set. For each of such local sets, we compute PCA and obtain explained variance ratios $\sigma_1 \leq \sigma_2 \leq \sigma_3$ such that $\sum_{i=1}^3 \sigma_i = 1$. Note that due to ordering it is sufficient to consider only $\sigma_2$ to decide whether the local set is near corner in a following way:
\begin{equation}
\label{eq:corner-det}
    \begin{split}
        \mathcal{N}_{\text{corner}} &= \{\mathcal{N}_i \, | \, \sigma_2^i > T_{\text{variance}}\} \\
        \mathcal{N}_{\text{curve}} &= \{\mathcal{N}_i \, | \, \sigma_2^i \leq T_{\text{variance}}\}.
    \end{split}
\end{equation}
Finally, we extract corner clusters by selecting a bounding box around the detected corner sets.

\begin{figure*}
\centering
\includegraphics[width=0.8\textwidth]{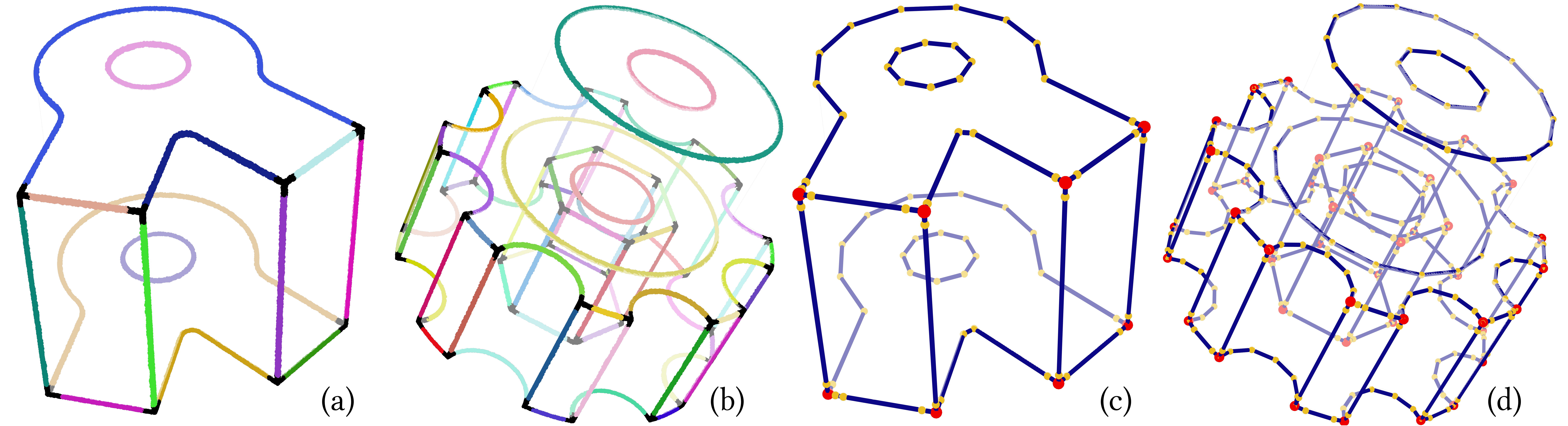}
\caption{Results of intermediate stages of wireframe extraction pipeline: (a), (b) -- examples of corner detection (black clusters) and segmentation procedure results; (c), (d) -- examples of the extracted topological graphs with corner points (red).}
\label{fig:seg-topo}
\end{figure*}

\subsubsection{Shape Curve Segmentation}
\label{sec:design:seg}
In order to retrieve the curve separation, we apply the knowledge of pairwise point sampling distance $r$, which is easy to estimate in practice. We eliminate corner clusters from $\mathcal{P}_{\text{sharp}}$ and get disconnected sets of points $\mathcal{P}_{\text{curves}}$ that substitute curves. The idea is to apply clustering techniques to separate them. To improve the separation, we construct dense $k$NN graphs from $\mathcal{P}_{\text{curves}}$: we connect points that are within $r$ sampling distance from each other. By construction, this guarantees that (1) all points inside one cluster are connected and (2) there are no connections between the separate clusters since we eliminated corner neighborhoods.

Once $k$NN graph is built, we detect connected components of such graph. Any connected component is a distinct curve cluster. We show the segmentation results in Fig.~\ref{fig:seg-topo}.

\subsubsection{Extraction of Topological Graph}
\label{sec:design:graph}
After the segmentation is done, we construct a topological graph fitted to $\mathcal{P}_{\text{sharp}}$. For this, we separately process each curve cluster. First, we detect endpoints for each curve by embedding it into a line. Consider points $p_i$; for each of them, we assign a linear ordering $\tau_i \in \mathbb{R}$ along the direction of maximal variance. The endpoints are the points, for which the value $\tau_i$ is minimal or maximal. 

Once this is finished, we initialize the polyline with two endpoints as nodes and a straight segment between them as an edge. Then, we recursively subdivide it. The potential breakpoints are obtained as:
\begin{equation}
\label{eq:split}
    p_{\text{split}} = \argmax_{p_i} \min_l \lvert \widehat d_i - \|p_i - \pi^l(p_i)\| \rvert,
\end{equation}
where $p_i$ is a point from the current curve cluster, $\pi^l(p_i)$ is a projection of $p_i$ onto the polyline segment $l$. Essentially, this formula computes a difference between the predicted distances $\widehat d_i$ and the empirical gap between a point and the nearest polyline segment. If this difference is greater than the threshold $T_{\text{split}}$, we assign $p_{\text{split}}$ as a new polyline node and split the corresponding segment in two. Additionally, we detect closed curves as curve clusters with no endpoints and process them separately. To create a polyline for a closed curve, we sample three points from the cluster and connect them to compose a triangle. After that, the subdivision process is similar.

The resulting polyline endpoints are then connected to the nearest corner cluster center, resulting in a finished topological graph $G(n,e)$ defined by the node positions $n$ and connections between them $e$. After this step is finished, we can define corners as graph nodes with more than two incident segments. Refer to Fig.~\ref{fig:seg-topo} for illustration.

\subsubsection{Spline Fitting}
\label{sec:design:splines}
Finally, the last stage is spline fitting and optimization. We identify path endpoints as graph nodes with degree not equal to 2. Subsequently, we partition the topological graph into shortest paths between such path endpoints, each path serving as a proxy to a curve. For a path $g_i = \left(n_i^1, n_i^2, \ldots n_i^{k_i}\right)$ represented as a sequence of graph nodes $n_i = \{n_i^j\}_{j=1}^{k_i}$ we get a set of nearest points $\mathcal{P}_i$, and compute projections $\pi^{g_i}(p_j), \, p_j \in \mathcal{P}_i$ and obtain values of parameters $u_i = \{u_i^j\}_{j=1}^{\lvert \mathcal{P}_i \rvert}$ as a cumulative sum of norms of $\pi^{g_i}(p_j)$ along the path $g_i$. Simultaneously, we compute knots $t_i = \{t_i^j\}_{j=1}^{k_i}$ as parameters of polyline nodes.

Fitting a spline results in a set of control points $c_i$ that define the exact spline curve. Once the spline is fitted, we can evaluate points $\mathcal{P}_i(c_i) = \gamma(u_i, \mathcal{P}_i, t_i, c_i)$ on the spline curve. These points, ideally, should be precisely as far away from point cloud points $\mathcal{P}_i$ as a distance field $\widehat d$ suggests. To enforce this property, we optimize over control points to shape the spline to the distance values:
\begin{equation}
\label{eq:spline-opt}
\begin{split}
    \sum_{j=1}^{\lvert \mathcal{P}_i \rvert} \left( \widehat d_i^j - \|p_i^j - \gamma(u_i^j, p_i^j, t_i, c) \| \right) ^2 \rightarrow \min_c,
\end{split}
\end{equation}
where $p_i^j \in \mathcal{P}_i$, $\widehat d_i^j$ is a corresponding distance value, and $\gamma(u_i^j, p_i^j, t_i, c)$ is a point corresponding to $p_i^j$ evaluated on the spline. Additionally, we impose constraints on the spline endpoints to match the polyline endpoints.

\begin{figure*}
\centering
\includegraphics[width=0.8\textwidth]{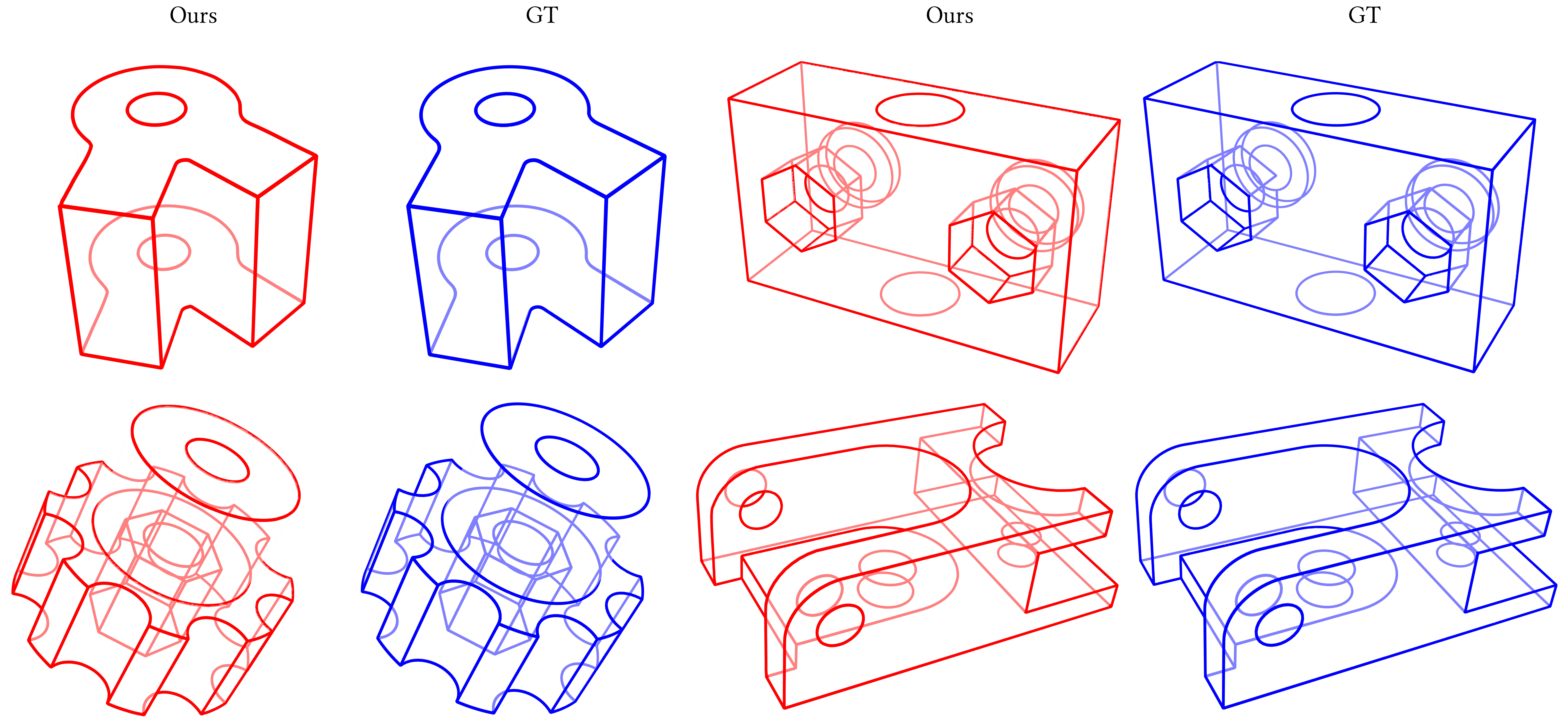}
\caption{Results of our wireframe extraction pipeline (red) compared to ground truth parametric curves (blue).}
\label{fig:splines}
\end{figure*}

The optimization problem and constraints are similar for the closed curves: endpoints of the spline should meet at the same point, and the tangents at the endpoint positions should be equal. For the results of spline fitting, refer to Fig.~\ref{fig:splines}.

\section{Experiments}
\label{sec:experiments}

\subsection{Dataset and Implementation Details}
We have selected a diverse set of 50 CAD models from the ABC dataset. When choosing these shapes, we focused on covering the diversity in such properties as variation in surface types (plane, cylinder, spline), feature curve types, sharpness angles. The selected meshes have been sampled following the DEF paper and then processed with a DEF neural network to obtain distance field predictions. 

Our wireframe extraction pipeline was implemented in Python. For graph subroutines (paths search, connected components separation), we employed networkx~\cite{hagberg2008exploring}. In addition, we used PCA implementation from scikit~\cite{pedregosa2011scikit} for corner and endpoint detection and KDTree nearest search and optimization techniques implemented in scipy~\cite{virtanen2020scipy} for nearest neighbors search, spline fitting, and optimization.

Here we discuss the choice of parameters. By design of sampling technique, the pairwise point distance $r$ is 0.02 on average; we choose to relate all parameters to this value. The sharpness threshold, which is used to obtain point cloud skeleton $\mathcal{P}_{\text{sharp}}$ is equal to $T_{\text{dist}} = 1.5r$. The choice of $T_{\text{dist}}$ is conditioned by the need to have a sufficient number of samples in the skeleton.  On the other hand, one needs to have thin lines to have robust corner detection and avoid false positives.

In the corner detector, one needs to select ball radius and variance threshold. The ball should be of sufficient size to capture geometry and should be greater than the line width $T_{\text{dist}}$. In our implementation, $R_{\text{corner}} = 4r$. We set $T_{\text{variance}} = 0.3$, this value for the second largest variance ratio indicates a sufficiently non-linear local set shape.

Finally, the choice of splitting threshold is $T_{\text{split}} = 4r$. We argue that this value should be again greater than $T_{\text{dist}}$ in order to prevent overfitting. However, we want the polyline controlled by this value to reflect the corresponding curve geometry accurately.

\subsection{Numerical Evaluation}
To assess the wireframe quality, we ran our pipeline on the selected point cloud shapes along with PIE-Net and computed several metrics to compare results to the ground truth parametric curves. 

We report two metrics commonly used to measure the distance between sets. First, we compute Chamfer distance defined as:
\begin{equation}
\label{eq:cd}
\begin{split}
    CD(X, Y) &= \frac{1}{2} \left(CD_{X \rightarrow Y} + CD_{Y \rightarrow X} \right) \\
    CD_{X \rightarrow Y} &= \frac{1}{n_x} \sum_{x\in X} \inf_{y \in Y} \|x - y \|.
\end{split}
\end{equation}
Chamfer distance approximates the continuous norm between two functions and reflects the average discrepancy in two sets of curves. We sampled the predicted wireframes and the ground truth set of curves into point sets and derived distances between the closest points to calculate it. Our pipeline resulted in a $CD$ value of 0.07. We remind that the average sampling distance is 0.02, which means that, on average, our approach is within 4 sampling distances to the real wireframe. PIE-Net output has order of magnitude larger error.

Secondly, we report Hausdorff distance defined as:
\begin{equation}
\label{eq:hd}
    HD(X,Y) = \max\{ \sup_{x \in X} \inf_{y \in Y} \| x - y \|, \,  \sup_{y \in Y} \inf_{x \in X} \| x - y \|\}.
\end{equation}

\begin{figure*}
\centering
\includegraphics[width=0.8\textwidth]{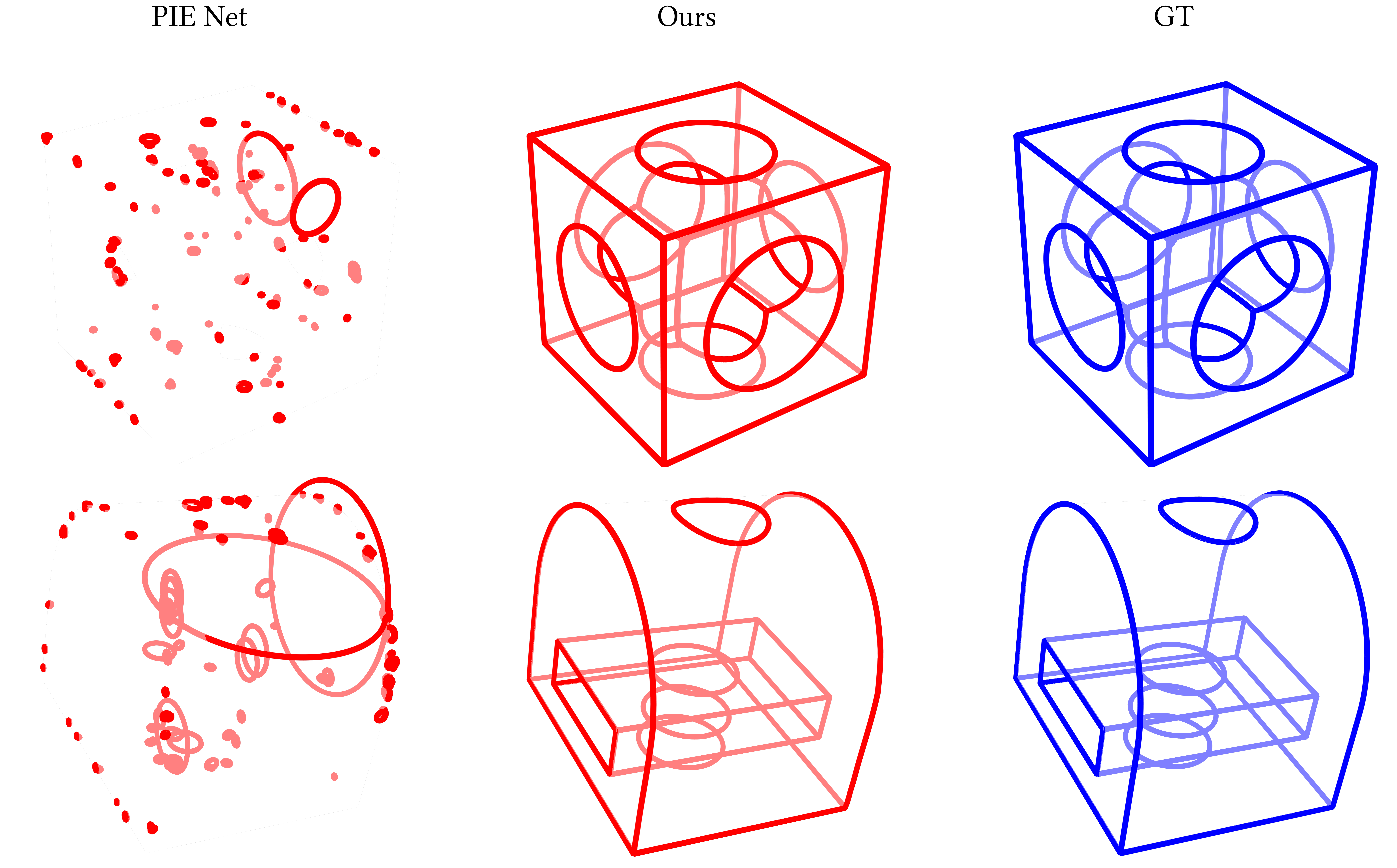}
\caption{Comparisons of our wireframe extraction pipeline with the results obtained with PIE-Net.}
\label{fig:comparisons}
\end{figure*}

Hausdorff distance measures the worst-case deviation between the curves. Again, the reported values suggest that our method has an error smaller than PIE-Net. We report the metrics in Tab.~\ref{tab:results} and demonstrate the comparisons in Fig.~\ref{fig:comparisons}. 

Not all of the shapes resulted in the final set of parametric curves. We report the percentage of the fail cases also in Tab.~\ref{tab:results}. Our method has a fail rate of 12\% of shapes. The numerical instabilities have caused most failures in spline fitting. In the case of PIE-Net, their failure rate is 66\%, and although the rest objects have outputted some set of curves, commonly, it is inaccurate and fails to fit the shape. Most likely, this issue is related to the limitations of the PIE-Net, which was trained on small shapes of 8,192 points. Overall, our method successfully extracts correct wireframes when PIE-Net fails to do so, even for relatively simple shapes.

\begin{table}[]
\centering
\begin{tabular}{@{}lccc@{}}
\toprule
\begin{tabular}[c]{@{}l@{}} Method \end{tabular} &
\begin{tabular}[c]{@{}c@{}} Chamfer distance \end{tabular} &
\begin{tabular}[c]{@{}c@{}} Hausdorff distance \end{tabular} &
\begin{tabular}[c]{@{}c@{}} \# fails, \% \end{tabular} \\
\cmidrule(r){1-4}
PIE-Net & 0.34 & 1.45 & 66.0\\ 
Ours & 0.07 & 0.6 & 12.0 \\
\bottomrule
\end{tabular}
\caption{Results of the numerical evaluation of our wireframe extraction pipeline and PIE-Net.}
\label{tab:results}
\end{table}

\subsection{Limitations}
Here we discuss the major limitations of our pipeline. Firstly, our method requires several parameters to be set, which can be overcome by introducing learnable stages into the approach. Our method highly depends on the predicted distance estimates, which results in faulty curve connections. This effect can be reduced by careful choice of parameters. Our evaluation was done of densely sampled clean points clouds, and we acknowledge that such data is rare in reality. However, our experiments suggest that, for accurate distance field predictions and careful parameter choice, the noise effect can be reduced, and our method would output a reasonable result. Evaluation on the real-world scan data is a future work. 

\section{Conclusion}
\label{sec:conclusion}
This paper presented a parametric wireframe extraction pipeline that processes per-point distance-to-feature estimates and consists of four major parts. The first stage is corner detection in point cloud skeleton, where we proposed a novel distance-based cornerness measure. In the second stage, we obtain curve segmentation. The third part is topological graph construction, which enables an accurate topological description. The last part is the extraction of parametric curves by fitting spline curves to paths in a topological graph.

Our approach can construct complex wireframes for shapes with high geometrical complexity. Each extracted curve, in general, corresponds to a separate smooth curve on the shape, which means our method does not overestimate the number of curves. The output may be easily processed in multiple CAD and graphical programs; one can query analytical differential quantities for curves and sample them with arbitrary resolution.

We have experimentally tested our method by running it on a large set of shapes and reported the quality of the fitted wireframes. We demonstrate that our approach is more robust and produces more meaningful results than the recent learning-based alternative. Besides having better quality measures, our pipeline outputs meaningful results in most cases.

Possible directions to improve our approach include adding learning-based modules. Specifically, corner detection and segmentation could vastly reduce the number of parameters needed and improve the robustness and generality of the approach. Also, we note that the real scan data evaluation of our method is a future work.

\bibliographystyle{ACM-Reference-Format}
\bibliography{99-bibliography}

\end{document}